\pdfoutput=1

\documentclass[11]{article}

\usepackage[final]{acl}
\usepackage{float}
\usepackage{times}
\usepackage{latexsym}

\usepackage[T1]{fontenc}

\usepackage[utf8]{inputenc}
\usepackage{makecell}
\usepackage{microtype}

\usepackage{inconsolata}

\usepackage{graphicx}
\usepackage{booktabs}
\usepackage{multirow}
\usepackage{amsmath}
\usepackage{comment}
\usepackage{xcolor}
\usepackage{pifont}
\newcommand{\cmark}{\ding{51}}
\newcommand{\xmark}{\ding{55}}
\title{LAURA: Knowledge Distillation for Interpretable Ambiguous Clause
Identification in Legal Contracts}

\author{
  Amrita Singh, 
  Aditya Joshi,  
  Jiaojiao Jiang,
  Hye-young Paik\\
  School of Computer Science and Engineering \\
  University of New South Wales (UNSW), Sydney \\
}

\begin{document}
\maketitle
\begin{abstract}
Legal contracts contain ambiguities that expose enterprises to financial and legal risks. Some ambiguities allow flexible interpretation without triggering disputes, while others lead to significant legal conflicts. This makes identification alone insufficient, and interpretable rationale analysis essential. We propose LAURA, a post-training framework for interpretable ambiguous clause identification. LAURA leverages knowledge distillation with an IRAC-Unlearning prompting technique to transfer knowledge from a teacher LLM to an open-weight student model (<=1B parameters), which is then trained using a joint objective combining classification and rationale generation losses. The framework supports both legal and non-legal stakeholders in making informed decisions about which ambiguities require further attention. Extensive experiments across 7 baselines and 7 open-weight models demonstrate that LAURA with Flan-T5 (250M) delivers state-of-the-art interpretability over all interpretable baselines while matching the identification performance of the best-performing opaque baseline.
\end{abstract}

\section{Introduction}
\label{section1}
Commercial enterprises manage numerous legal contracts, requiring thorough reviews to prevent serious penalties from breaches \cite{singh2024data}. Reviews are challenging due to complex legal language (often called \emph{Legalese}) and inherent ambiguities that can lead to conflicting interpretations in litigation \cite{martinez2024even, martinez2022poor, singh2025survey, xu2022conreader, barale2025lextime}. In contract analysis, ambiguity arises when a clause admits multiple reasonable interpretations or constructions \cite{garner2014blacks}. Legal text ambiguity encompasses six categories identified by \citet{massey2014identifying}, whose definitions and examples are provided in Appendix \ref{AppendixDef}.
\citet{singhal2024generating} introduce the first and only publicly available dataset on contract ambiguity identification and address ambiguous clause identification by generating clarification questions for each clause using dense retrieval-based methods. However, interpretability is absent from their approach, a limitation the authors themselves acknowledge. Our work focuses on interpretable ambiguous clause identification in legal contracts. The objective is to accompany each classification label (ambiguous or not ambiguous) with a textual rationale that: (i) extracts keywords or phrases and evaluates if they introduce ambiguity; (ii) if ambiguous, explains how the identified terms create conflicting interpretations; and (iii) highlights the resulting legal consequences, where applicable. This is crucial because not all ambiguities are equally consequential \cite{li2017corpus}; some allow flexible interpretation without triggering disputes, while others lead to significant legal conflicts. Since identification alone cannot capture this distinction, interpretable rationale analysis is essential.

We propose \textbf{LAURA} (\textbf{L}egal \textbf{A}mbiguity \textbf{U}nderstanding via \textbf{R}ationale \textbf{A}nalysis), a novel framework for interpretable ambiguous clause identification that utilises knowledge distillation (KD). Vanilla KD utilises a teacher-student model pair where distillation transfers logits or hidden states \cite{mansouriancomprehensive}. In contrast, LAURA distills legally grounded rationales via a novel IRAC-Unlearning prompting technique, eliciting structured and faithful rationales from the teacher model and transferring this knowledge to the student model. The student model is trained with a joint objective that combines classification and rationale generation losses, enabling it to predict ambiguity labels and produce human-readable rationales. Evaluated on the only publicly available ambiguous clause identification dataset \cite{singhal2024generating}, LAURA achieves state-of-the-art interpretability without dropping identification performance compared to the best baseline, across a rigorous evaluation spanning diverse configurations (7 baselines, 7 models, and 5 LAURA variants), complemented by qualitative rationale assessment across three dimensions (correctness, completeness, and conciseness) and detailed classification error analysis.
\begin{figure*}[ht!]
\centering
\includegraphics[width=0.76\textwidth]{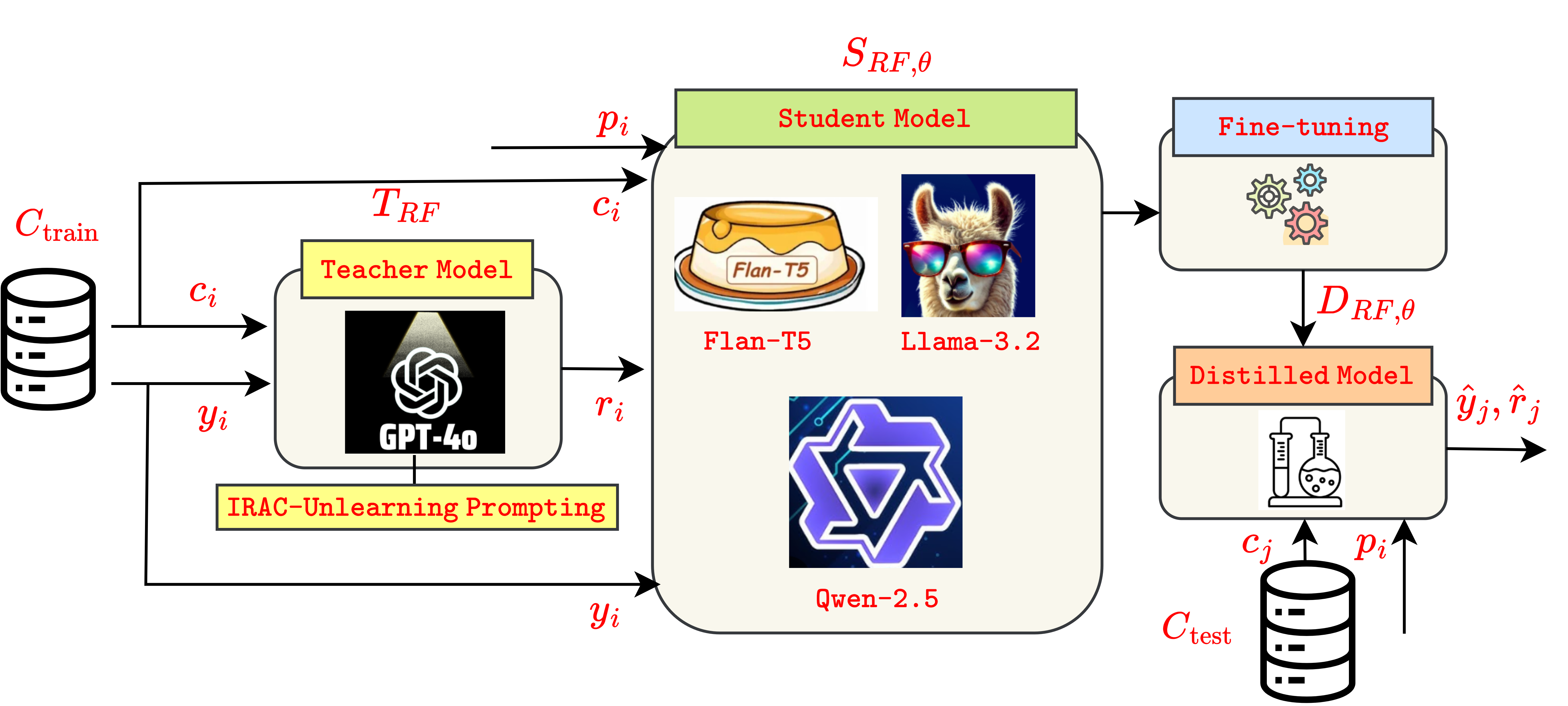}
\caption{Architecture of the LAURA Framework}
\label{Figure2}
\end{figure*}

To the best of our knowledge, and supported by a recent survey \cite{singh2025survey}, \textbf{ours is the first work on interpretable ambiguous clause identification}, one of the challenging tasks in legal contract review. An additional advantage of LAURA is in terms of scalability and privacy. Reliance on commercial LLMs entails dependence on cloud-based APIs, which is costly and not scalable  when processing contracts clause by clause across documents spanning hundreds of pages. This also introduces privacy risks, which are especially acute for sensitive legal contracts \cite{wang2024comprehensive, van2024survey, subramanian2025small, lu2024small}. LAURA uses open-weight models and does not depend on commercial APIs at inference time. To the best of our knowledge, no prior work addresses interpretability in ambiguous clause identification using open-weight models (<=1B), despite their reductions in training cost, inference latency, and energy consumption.

\section{Framework of LAURA}
The architecture of the \textbf{LAURA} (\textbf{L}egal \textbf{A}mbiguity \textbf{U}nderstanding via \textbf{R}ationale \textbf{A}nalysis) framework, is illustrated in Figure \ref{Figure2}. The clause set $C$ is partitioned into a training set $C_{\text{train}}$ and a test set 
$C_{\text{test}}$, where each training clause $c_i \in C_{\text{train}}$ carries a binary 
label $y_i \in \{0, 1\}$: $y_i = 1$ denotes an ambiguous clause and $y_i = 0$ denotes an 
not ambiguous clause. A teacher model $T_{\text{RF}}$ (GPT-4o) generates a rationale set 
$R = \{r_1, r_2, \dots, r_N\}$, where each rationale $r_i = T_{\text{RF}}(c_i, y_i)$ is 
produced conditioned on both the clause and its ground-truth label, ensuring rationale correctness and preserving full dataset integrity \cite{rejithkumar2025nice}. This is particularly critical as only one limited-size annotated dataset exists for this task, with no rationales \cite{singhal2024generating}. To elicit faithful rationales and mitigate hallucination, we employ multiple prompting strategies: Chain-of-Thought (CoT) \cite{wei2022chain}, Few-shot, Contrastive \cite{jung2025courtroom}, and SC-Unlearning prompting \cite{kamoi2024can, zhang2025understanding}. Additionally, inspired by the IRAC legal reasoning framework \cite{burton2017think, yu2025benchmarking}, we introduce a novel IRAC-Unlearning prompting technique to elicit faithful rationales from $T_{\text{RF}}$ in LAURA. Grounded in the canonical legal reasoning methodology, IRAC-Unlearning scaffolds rationale generation across four structured stages: Issue, Rule, Application, and Conclusion, mirroring the interpretive process lawyers employ when analyzing contractual ambiguity, augmented with an unlearning signal ensuring that elicited rationales are both domain-faithful and transferable to the student model. The prompts used to elicit the rationale from the teacher model are provided in Appendix \ref{AppendixP}.

The student model $S_{\text{RF},\theta}$ is then fine-tuned to receive a concatenated input 
of prompt $p_i$ and clause $c_i$, and to produce the joint output $(\hat{y}_i, \hat{r}_i)$, 
a predicted label followed by a rationale token sequence. Training minimizes the introduced combined 
classification and rationale generation loss:
\begin{align}
\mathcal{L}_{\text{RF}}(\theta) &= \frac{1}{|C_{\text{train}}|} \sum_{i=1}^{|C_{\text{train}}|} 
\Big[ \ell_{\text{cls}}(\hat{y}_i, y_i) + \lambda\,\ell_{\text{rat}}(\hat{r}_i, r_i) \Big],
\end{align}
where $\ell_{\text{cls}}$ is the binary cross-entropy (BCE) classification loss, 
$\ell_{\text{rat}}$ is the sequence-level cross-entropy rationale loss measuring alignment 
between the predicted rationale $\hat{r}_i$ and the teacher rationale $r_i$, and $\lambda$ 
is a weighting hyperparameter (set to $1$ in all experiments, as ambiguous 
clause identification and interpretability are equally important for the task). 
\begin{align}
\mathcal{L}_{\text{RF}}(\theta) &= \frac{1}{|C_{\text{train}}|} \sum_{i=1}^{|C_{\text{train}}|} \Bigg[
 -y_i \log(\hat{y}_i) \nonumber\\
&\quad - (1 - y_i) \log(1 - \hat{y}_i) \nonumber\\
&\quad + \lambda \left( - \sum_{t=1}^{T} r_{i,t} \log \hat{r}_{i,t} \right)
\Bigg],
\end{align}
where $(\hat{y}_i, \hat{r}_i) = S_{\text{RF},\theta}(p_i, c_i)$ and $T$ denotes the rationale 
sequence length. Upon completion of fine-tuning, the student becomes the distilled model 
$D_{\text{RF},\theta}$, which is evaluated on the held-out test set as:
\begin{equation}
(\hat{y}_j, \hat{r}_j) = D_{\text{RF},\theta}(p_i, c_j), \quad \forall\, c_j \in C_{\text{test}}.
\end{equation}

LAURA simultaneously identifies ambiguous clauses and generates human-readable, 
faithful rationales in a single forward pass. Since the rationale is produced as a token 
sequence immediately following the label prediction, the framework requires sequence-generating 
architectures, either encoder-decoder or decoder-only, capable of autoregressive generation.

\section{Experiment Setup}
We evaluate LAURA on the only publicly available contract ambiguity dataset, introduced by \citet{singhal2024generating}. This dataset contains $1,000$ clauses  across 25 contract types, annotated with a boolean ambiguity label: $524$ ambiguous (if they contain vagueness, incompleteness, or referential ambiguity) and $476$ not ambiguous. We use an 80:20 train-test split. No rationales are available in the dataset.  We consider seven baselines. The first is the majority baseline, which predicts the ambiguous class for all instances. The remaining six span two categories. Three prompt-based baselines are evaluated: \emph{Direct} \cite{kojima2022large}, \emph{CoT} \cite{wei2022chain}, and \emph{Few-Shot} \cite{brown2020language}, which use an LLM (GPT-4o) to classify clauses and generate rationales. Three fine-tuning-based baselines are also compared: (i) \emph{SFT} \cite{ouyang2022training} trains on clause-label pairs; (ii) \emph{IFT} \cite{weifinetuned} takes an instruction and clause as input and trains to predict the label; (iii) \emph{PPI} takes the prediction from IFT and prompts the base model with IRAC-Unlearning to generate rationales from the clause and predicted label, such that IFT and PPI share the same quantitative metrics, but PPI also generates rationales without any additional training. All baseline prompt templates, wherever applicable, are provided in Appendix \ref{AppendixA}. Models and hyperparameters in Appendix \ref{AppendixB}. We report binary precision, binary recall, and binary F1-score for the ambiguous class, along with accuracy, following the evaluation setting of \citet{singhal2024generating}. We emphasize accuracy as the primary metric for overall system performance given the balanced dataset, and binary F1-score to measure how well the model detects ambiguous clauses. We further perform classification error analysis and qualitative evaluation of rationales across three dimensions: correctness, completeness, and conciseness, with their definitions provided in Table \ref{tab1} in Appendix \ref{Appendix00} due to space constraints.
\begin{table}[ht!]
  \centering
  \begin{tabular}{@{}l@{\hspace{4pt}}l@{\hspace{4pt}}l@{\hspace{4pt}}l@{\hspace{4pt}}l@{\hspace{4pt}}l@{\hspace{4pt}}l@{}}
    \toprule
    App. & Models & P & R & F1 & A & I\\
    \midrule
    Majority & -- & 0.52 & 1.0 & \textbf{0.68} & \textbf{0.52} & \xmark \\ 
    \midrule
    Direct & GPT-4o & 0.53 & 0.97 & \textbf{0.69} & \textbf{0.54} & \cmark \\ 
    \midrule
    CoT & GPT-4o & 0.53 & 0.98 & \textbf{0.69} & \textbf{0.53} & \cmark \\ 
    \midrule
    3-Shot & GPT-4o & 0.69 & 0.32 & \textbf{0.44} & \textbf{0.57} & \cmark \\ 
    5-Shot & GPT-4o & 0.63 & 0.30 & 0.40 & 0.54 & \cmark \\
    \midrule
    SFT & BERT & 0.71 & 0.60 & 0.65 & 0.66 & \xmark \\
    & RoBERTa & 0.72 & 0.68 & 0.70 & \textbf{0.69} & \xmark \\
    & \makecell[l]{Legal-\\BERT} & 0.61 & 0.82 & 0.70 & 0.64 & \xmark \\
    & \makecell[l]{Contracts-\\BERT} & 0.69 & 0.70 & 0.70 & 0.68 & \xmark \\
    & Flan-T5 & 0.64 & 0.77 & 0.70 & 0.66 & \xmark \\
    & Qwen-2.5 & 0.59 & 0.87 & \textbf{0.71} & 0.62 & \xmark \\
    & Llama-3.2 & 0.70 & 0.65 & 0.67 & 0.67 & \xmark \\
    \midrule
    IFT/PPI & Flan-T5 & 0.55 & 0.97 & 0.70 & 0.56 & \xmark/\cmark \\
    & Qwen-2.5 & 0.61 & 0.85 & \textbf{0.71} & 0.63 & \xmark/\cmark \\
    & Llama-3.2 & 0.67 & 0.72 & 0.70 & \textbf{0.67} & \xmark/\cmark \\ 
    \midrule
    LAURA & Flan-T5 & 0.62 & 0.46 & 0.53 & 0.57 & \cmark \\
    + & Qwen-2.5 & 0.69 & 0.67 & \textbf{0.68} & \textbf{0.67} & \cmark \\
    CoT & Llama-3.2 & 0.62 & 0.67 & 0.64 & 0.61 & \cmark \\ 
    \midrule
    LAURA & Flan-T5 & 0.70 & 0.47 & 0.56 & 0.62 & \cmark \\
    + & Qwen-2.5 & 0.63 & 0.71 & \textbf{0.67} & \textbf{0.64} & \cmark \\
    Few-Shot & Llama-3.2 & 0.66 & 0.55 & 0.60 & 0.61 & \cmark \\
    \midrule
    LAURA & Flan-T5 & 0.71 & 0.57 & 0.63 & \textbf{0.66} & \cmark \\
    + & Qwen-2.5 & 0.64 & 0.73 & 0.68 & 0.65 & \cmark \\
    Contrastive & Llama-3.2 & 0.58 & 0.87 & \textbf{0.69} & 0.60 & \cmark \\
    \midrule
    LAURA & Flan-T5 & 0.69 & 0.65 & 0.67 & 0.67 & \cmark \\
    + & Qwen-2.5 & 0.64 & 0.66 & 0.65 & 0.63 & \cmark \\
    SC-Unlearning & Llama-3.2 & 0.67 & 0.71 & \textbf{0.69} & \textbf{0.67} & \cmark \\ 
    \midrule
    LAURA & Flan-T5 & 0.71 & 0.68 & \textbf{0.70} & \textbf{0.69} & \cmark \\
    & Qwen-2.5 & 0.63 & 0.45 & 0.54 & 0.58 & \cmark \\
    & Llama-3.2 & 0.69 & 0.50 & 0.58 & 0.63 & \cmark \\
    \bottomrule
  \end{tabular}
  \caption{Evaluation for approach (app.) and model combinations on the test set. Highest F1 and Accuracy per approach is in bold. P, R, F1, A, I: Binary Precision, Binary Recall, Binary F1-score, Accuracy, and Interpretability (indicated by \cmark and \xmark). \#params in GPT-4o: 1.8T, RoBERTa: 125M, BERT, Legal-BERT \& Contracts-BERT: 110M, Flan-T5: 250M, Qwen-2.5: 0.5B, Llama-3.2: 1B.}
  \label{tab2}
\end{table}

\section{Results and Analysis}
\label{result}
\begin{figure}[ht!]
  \centering
  \includegraphics[width=0.8\linewidth]{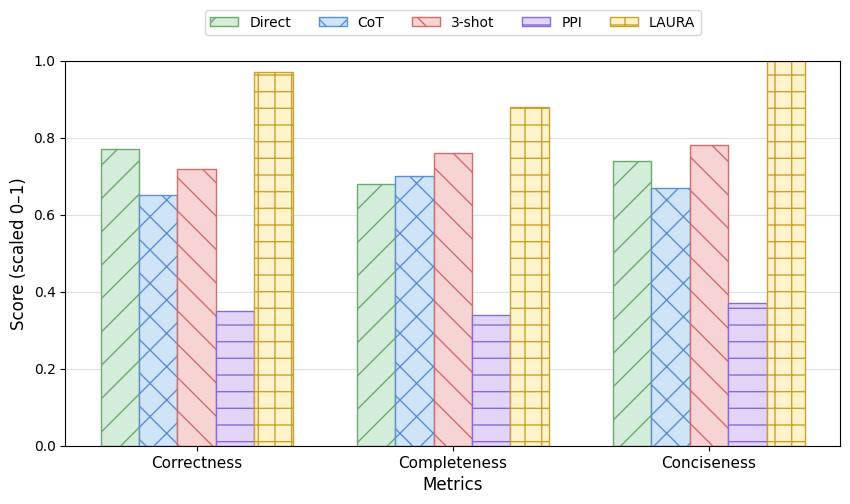}
  \caption{LAURA vs. Baseline Rationale Quality}
  \label{fig2}
\end{figure}

\begin{figure}[ht!]
  \centering
  \includegraphics[width=0.8\linewidth]{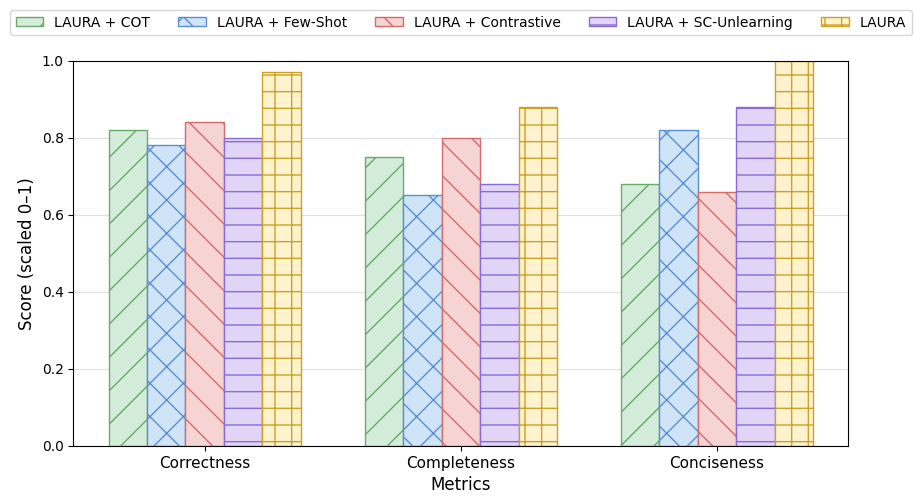}
  \caption{LAURA vs. Variant Rationale Quality}
  \label{fig3}
\end{figure}

Table \ref{tab2} presents the identification performance of LAURA and its variants across different prompting techniques, baselines, and models. Overall, LAURA achieves state-of-the-art interpretability (Figures \ref{fig2} and \ref{fig3}) over all interpretable baselines while maintaining overall system performance (Table \ref{tab2}), achieving a binary F1-score of 0.70 and an accuracy of 0.69 using Flan-T5 (250M parameters). The majority baseline achieves a binary F1-score of 0.68 and accuracy of 0.52 by predicting every clause as ambiguous, serving as the lower bound for evaluation. Direct and CoT both achieve a binary F1 of 0.69 and accuracy of 0.53 and 0.54 respectively, barely above the majority baseline. Both have very high recall (0.97 and 0.98) but very low precision (0.53), indicating they classify almost every clause as ambiguous, similar to the majority baseline. Few-Shot prompting significantly drops F1 to 0.44 and 0.40, the lowest among all approaches. Precision improves (0.69 and 0.63) but recall collapses (0.32 and 0.30), meaning the model misses many ambiguous clauses. This suggests that few-shot examples push the model toward predicting not ambiguous, hurting overall performance, achieving accuracy of 0.57 for 3-Shot and 0.54 for 5-Shot. \textit{These results indicate that prompt-based baselines are not effective for ambiguous clause identification, as they demonstrate poor identification performance.} 

SFT performs comparatively better than prompting-based approaches with binary F1 ranging from 0.65 to 0.71 and accuracy from 0.62 to 0.69. Qwen-2.5 achieves the best binary F1 of 0.71 while RoBERTa achieves the best accuracy of 0.69. IFT/PPI achieves similar binary F1 to SFT (0.70 to 0.71) but with higher recall and lower precision, indicating over-prediction of ambiguous clauses. Qwen-2.5 performs best with binary F1 of 0.71 and Llama-3.2 achieves the best accuracy of 0.67. However, no single model across SFT, IFT, and PPI achieves strong performance on both metrics simultaneously except RoBERTa under SFT, which achieves binary F1 of 0.70 and accuracy of 0.69. \textit{This indicates that correctly identifying ambiguous clauses while avoiding misclassification of not-ambiguous ones is a difficult task.} In contrast, the proposed LAURA using Flan-T5 balances both binary F1-score (0.70) and accuracy (0.69), matching the performance of the best-performing opaque baseline, RoBERTa under SFT. 

To evaluate interpretability, we sample 50 correctly predicted clauses across 9 approaches to evaluate rationale quality, and 50 incorrectly predicted clauses from LAURA for classification error analysis. This covers 450 rationales annotated across three criteria resulting in 1,350 annotations, plus 50 additional annotations for classification error analysis, totalling 1,400 annotations completed over 3 person-days. Two annotators perform the annotation, one with over four years of Legal NLP experience specialising in contract annotation and requirement implementation, and the other a final-year law student specialising in contract law. We compute IAA on a 90-example sample with 10 examples drawn from each approach, achieving Cohen's $\kappa$ = 0.93 (almost perfect agreement \cite{landis1977measurement}), with disagreements resolved through consensus prior to qualitative analysis, after which remaining annotations are divided equally between the two annotators. Rationale quality is evaluated using a 3-point Likert scale across three criteria: Correctness (Cor), Completeness (Com), and Conciseness (Con), as defined in Table \ref{tab1}, where 1 indicates the lowest and 3 the highest quality, normalised to a 0-1 scale. Results across interpretable approaches with the best-performing model are presented in Figures \ref{fig2} and \ref{fig3}.
LAURA achieves the highest scores in correctness and conciseness across all baselines and variants. Direct, CoT, and 3-Shot perform moderately but fall well below LAURA, indicating that prompting LLMs without fine-tuning is insufficient for high-quality rationale generation for this task. While LAURA occasionally misses minor details reflected in lower completeness, it outperforms all baselines and variants across all three criteria. Among variants, LAURA+Few-Shot scores lowest on completeness suggesting that few-shot examples push the model to be too brief and miss key details, while LAURA+CoT and LAURA+Contrastive score lower on conciseness due to longer reasoning. PPI shows the lowest rationale quality as it relies solely on inference-time prompting without training, causing open-weight models (<=1B) to hallucinate. This highlights that training the student model is essential for better rationale generation. The 50 classification errors are of the following types (with frequency in brackets): Vagueness (19), incompleteness (2), referential (4) and not ambiguous (25). Most errors are due to vagueness of ambiguous clauses and misclassification of non-ambiguous clauses. Terms such as "reasonable", "applicable", or "good faith" are context-dependent: they can be precise in some contexts but vague in others, making consistent distinction challenging for the model. Next, the model over-flags standard legal phrases, cross-references to other sections, and redacted content as ambiguous, indicating that complex syntactic structures and dense legal terminology mislead the model into predicting ambiguity where none exists (LAURA outputs are presented in Appendix \ref{AppendixC} and additional analysis is provided in Appendix \ref{AppendixF}).

\section{Conclusion}
We present the first systematic evaluation of interpretable ambiguous clause identification using open-weight models (<=1B). We propose LAURA, a post-training framework for joint ambiguous clause identification and rationale generation. Experiments across diverse configurations (7 baselines, 7 models, and 5 LAURA variants) demonstrate that LAURA with Flan-T5 achieves state-of-the-art interpretability while maintaining identification performance, outperforming all interpretable baselines in rationale correctness and conciseness. Our results show that prompt-based approaches alone are insufficient and that task-specific model training is essential for correct and reliable interpretation. Error analysis further reveals that context-dependent vague terms and clause-level processing of dense legal language remain key challenges, highlighting promising directions for future research in interpretable ambiguous clause identification

\section*{Limitations}
Our work has several limitations. First, only one publicly available dataset exists for contract ambiguity, and it is relatively small and exclusively in English, limiting the scope of our experiments. Second, LAURA is evaluated only on legal contracts, and while the dataset is diverse in contract types, it may or may not generalise to other legal text types such as statutes, court cases, or legal opinions, as legal language and its interpretation vary significantly across genres. Third, LAURA requires the model to produce a label followed immediately by a rationale token sequence in a single forward pass, which is inherently a sequence-to-sequence task requiring both input understanding for classification and structured output generation for rationale generation. Flan-T5's encoder-decoder architecture is well suited for this, while Qwen-2.5 and Llama-3.2 as decoder-only models treat the entire input-output as one left-to-right sequence, making it harder to jointly optimise classification and rationale generation at small parameter counts. Fourth, ambiguity identification and interpretation are performed at the clause level, limiting access to broader document context. This makes it more difficult for LAURA to resolve cross-references, referential ambiguities, and context-dependent terms such as `reasonable', `applicable', and `good faith', which may be precise in some contexts but ambiguous in others. However, LAURA still performs well at the clause level, demonstrating its effectiveness despite this limitation. Fifth, LAURA occasionally omits details in its rationales and makes identification errors, as shown in Figures \ref{fig2} and \ref{fig3} and Table \ref{tab1}. These limitations reflect the challenges of jointly performing legal interpretation and ambiguity identification with open-weight models (<=1B). Nevertheless, this work represents a first systematic attempt at interpretable ambiguous clause identification using open-weight models (<=1B), validated through rigorous evaluation. These limitations open new research directions. Constructing larger multilingual contract ambiguity datasets would enable studying how ambiguity varies across languages and jurisdictions. Incorporating document-level context beyond the clause level could help resolve cross-references and referential ambiguities. Finally, exploring RLHF or human-LLM collaboration could improve rationale completeness and identification performance, advancing trustworthy legal contract review at scale.
\section*{Ethical Considerations}
This research uses a publicly available dataset to identify ambiguous clauses at an interpretable level, all of which contain contract clauses without personal data, sourced from CUAD, which is derived from public U.S. SEC EDGAR filings. However, potential ethical risks arise, including adversarial exploitation, model manipulation, and the introduction of unintended biases. Techniques such as LAURA could be misused to deliberately insert ambiguities into contractual clauses through prediction and rationale generation. Malicious actors might exploit these techniques to manipulate contracts in their favor, undermining legal integrity. Inaccuracies in prediction and rationale generation may also lead to unintended consequences, such as misinterpreting contractual terms, with significant legal or financial ramifications. Another ethical risk involves automation bias, or over-reliance on model outputs, where non-legal stakeholders or non-experts may assume results are always correct or unbiased. This can reduce critical oversight, particularly when models fail to capture clause meaning, intent, or subtle nuances essential for interpretation. Therefore, while these approaches support legal professionals, they are not substitutes for legal expertise and must be applied with caution and oversight.
\bibliography{custom}

\appendix
\section{Types of ambiguity in legal text: definitions and examples}
\label{AppendixDef}
These are the six types of ambiguities present in legal text:

\begin{itemize}
\item Lexical: A word or phrase that has multiple valid interpretations.
\item Syntactic: A sequence of words that can be grammatically interpreted in multiple valid ways, regardless of context.
\item Semantic: A sentence that can be interpreted in more than one way within its given context.
\item Vagueness: A sentence that contains phrases allowing for borderline cases, leading to unclear boundaries of meaning.
\item Incompleteness: A grammatically correct sentence that lacks enough information for a single, clear interpretation.
\item Referential: A word or phrase in a sentence lacks a clear reference, leading to confusion about its meaning.
\end{itemize}
Table \ref{Table3i} shows examples of various ambiguities in legal text. Identifying ambiguity without a rationale is challenging, even when clauses are known to be ambiguous. When it is unknown whether a clause is ambiguous or not, the rationale plays a crucial role in understanding and interpreting it.
\begin{table}[ht!]
\centering
\begin{tabular}
{p{2cm} p{4.8cm}}
\toprule
\textbf{Ambiguity Types} & \textbf{Examples} \\ \midrule
Lexical &  Enable a user
to electronically record, modify, and retrieve a patient’s active medication list as well as medication history for longitudinal
care.\\ 
Syntactic & Supplier will request Distributor immediately to assign its rights in the property.  \\
Semantic &  The contractor must deliver the project on time. \\
Vagueness &  Data Processor shall keep Personal Information sufficiently isolated from other data on the server in an appropriate manner to prevent it from being misused.\\
Incompleteness & The Service Provider shall notify the Customer in writing of any changes to the Service Level Agreement. \\
Referential & The Service Provider shall provide support to the Customer according to the applicable provisions in the Agreement.  \\
\bottomrule
\end{tabular}
\caption{Examples of Ambiguities in Legal Text \cite{massey2014identifying,singhal2024generating}.}
\label{Table3i}
\end{table}

\section{Rationale Evaluation Criteria}
The evaluation of rationales across three dimensions, correctness, completeness, and conciseness, and their definitions are provided in Table \ref{tab1}.
\label{Appendix00}
\begin{table*}[ht!]
\centering
\begin{tabular}{p{4.5cm}p{4.5cm}p{4.5cm}}
\toprule
\textcolor{red}{\textbf{Correctness}} & \textcolor{red}{\textbf{Completeness}} & \textcolor{red}{\textbf{Conciseness}} \\ \midrule
\textbf{Cor1: Factually incorrect:} The rationale does not accurately represent the prediction. There are significant factual errors or contradictions within the rationale that misalign with the predicted label. &
\textbf{Com1: Misses key points or explanations:} The rationale omits essential aspects or critical explanations related to the predicted label, making it incomplete and inadequate. &
\textbf{Con1: Overtly verbose and hard to follow:} The rationale is excessively long or includes irrelevant information, making it difficult to follow and understand the reasoning to the predicted label. \\
\textbf{Cor2: Mostly correct with minor misinterpretation:} The rationale does not contradict the predicted label, but contains minor misinterpretations that may lead to incorrect or misleading details, which do not adequately support the prediction. &
\textbf{Com2: Covers important aspects but omits details:} The rationale addresses the key points related to the predicted label but leaves out minor details, resulting in an incomplete explanation. &
\textbf{Con2: Contains unnecessary elaboration:} The rationale includes unnecessary elaboration or details extraneous to the understanding of the predicted label. This makes the rationale less concise but relatively clear. \\
\textbf{Cor3: Acceptable:} The rationale is fully aligned with the predicted outcome, providing a clear and accurate explanation for the prediction, and it faithfully supports the model's prediction. &
\textbf{Com3: Acceptable:} The rationale fully covers all important aspects of the predicted label, providing a complete and detailed explanation with all essential information. &
\textbf{Con3: Acceptable:} The rationale is succinct, providing relevant information without unnecessary elaboration. It effectively supports the predicted label without being overly verbose or lacking in clarity. \\
\bottomrule
\end{tabular}
\caption{Qualitative Metrics for Rationale Analysis}
\label{tab1}
\end{table*}

\section{Prompt used to elicit the rationale from the teacher model (GPT-4o)}
\label{AppendixP}
In this section, we discuss the prompting techniques used to elicit rationales from the teacher model (GPT-4o): Chain-of-Thought (CoT) \cite{wei2022chain}, Few-Shot, Contrastive \cite{jung2025courtroom}, SC-Unlearning \cite{kamoi2024can, zhang2025understanding}, and our novel IRAC-Unlearning prompting, described in the following section.

\subsection{Chain-of-Thought (CoT)}
\textbf{Instruction:} You are an expert in legal contract analysis. Given the following contract clause and its label, think step by step to provide a short, concise, and precise rationale.\\
\textbf{Input:}\\
\emph{Contract Clause:} []\\
\emph{Label:} []\\
\textbf{Output:}\\
Step 1: Identify keywords or phrases that are relevant to the classification and evaluate if they introduce ambiguity\\
Step 2: If ambiguous, explain how the identified terms create conflicting interpretations; if not ambiguous, explain why the clause has a single clear interpretation\\
Step 3: If ambiguous, highlight the resulting legal consequences where applicable\\
\emph{Rationale:} [rationale behind the classified label]

\subsection{Few-Shot}
\textbf{Instruction:} You are an expert in legal contract analysis. Given the following contract clause and its label, provide a short, concise, and precise rationale.\\
\textbf{Definitions:}\\
1. \textbf{Ambiguous:} A clause that can be reasonably interpreted or constructed in more than one way, containing vagueness, incompleteness, or referential ambiguity.\
\begin{itemize}
\item \textbf{Vagueness:} A clause that contains phrases allowing for borderline cases, leading to unclear boundaries of meaning.
\item \textbf{Incompleteness:} A grammatically correct clause that lacks enough information for a single, clear interpretation.
\item \textbf{Referential:} A word or phrase in a clause lacks a clear reference, leading to confusion about its meaning.
\end{itemize}
2. \textbf{Not Ambiguous:} A clause that has a single clear and Not Ambiguous interpretation.\\
\textbf{Example1:}\\
\emph{Contract Clause:} []\\
\emph{Label:} Ambiguous\\
\emph{Rationale:} []\\
\textbf{Example2:}\\
\emph{Contract Clause:} []\\
\emph{Label:} Not Ambiguous\\
\emph{Rationale:} []\\
\textbf{NOW YOUR TURN}\\
\textbf{Input:}\\
\emph{Contract Clause:} []\\
\emph{Label:} []\\
\textbf{Output:}\\
\emph{Rationale:} [rationale behind the classified label]
\subsection{Contrastive}
You are a legal contract expert. Given the following contract clause and its label, internally perform the following steps (do NOT output them):\\
1. Generate reasoning that supports the given label.\\
2. Generate a counter-argument for the opposite label.\\
3. Explain why the supporting reasoning outweighs the counter-argument.\\\\
Output ONLY the final rationale in this exact format:\\
This clause is ambiguous. [your rationale here]\\
OR\\
This clause is not ambiguous. [your rationale here]\\\\
Rules:\\
- No brackets, no labels, no headings, no Rationale: prefix\\
- Short, concise, and precise\\
- Must start with This clause is ambiguous. OR This clause is not ambiguous.\\
\textit{Clause:} []\\
\textit{Label:} []\\
\subsection{Issue-Rule-Application-Conclusion (IRAC)-Unlearning}
\label{IRAC}
\textbf{Role Setting:} You are a seasoned legal contract expert, proficient in contract law and highly familiar with legal standards for interpreting contractual language. Internally perform a full IRAC analysis (do NOT output any of these steps):\\\\
\textbf{I - ISSUE}\\
Identify the core interpretive question:\\
- What specific term(s) or phrase(s) in the clause are potentially unclear or legally uncertain?\\
- What is at stake if the clause is misinterpreted?\\\\
\textbf{R - RULE}\\ 
Apply the relevant legal standards:\\
- Contra proferentem: ambiguous terms are construed against the drafter.\\
- Plain meaning rule: words are given their ordinary meaning unless defined otherwise.\\
- Reasonable person standard: would a reasonable party understand this clause the same way?\\
- Are key terms defined, measurable, and enforceable?\\\\
\textbf{A - APPLICATION}\\
Apply the rules to the clause:\\
- Does the clause satisfy the plain meaning rule?\\
- Would a reasonable person interpret this clause consistently?\\
- Do any terms trigger contra proferentem concerns?\\
- Cross-check findings against the given label.\\\\
IF your analysis matches the label → confirm rationale.\\
IF your analysis contradicts the label → identify the gap, unlearn your assumption, and re-examine through the lens of the given label.\\\\
\textbf{C - CONCLUSION}\\
Output ONLY the final rationale in this exact format:\\
This clause is ambiguous. [your rationale here]\\
OR\\
This clause is not ambiguous. [your rationale here]\\\\
Rules:\\
- No brackets, headings, or Rationale: prefix\\
- Cite the specific term(s) driving the classification\\
- Short, concise, and precise\\
\textit{Clause:} []\\
\textit{Label:} []
\subsection{Self Correction (SC)-Unlearning}
You are a legal contract expert. Given the following contract clause and its label, follow these steps silently (do NOT output them):\\\\
\textbf{PHASE 1: UNBIASED ANALYSIS} \\
{[STEP 1 - FREE REASONING]:}\\
Read the clause WITHOUT knowing the label.\\
Generate an initial rationale based purely on the clause text.\\
Assign your own predicted label: Ambiguous OR Not Ambiguous\\\\
\textbf{PHASE 2: LABEL ALIGNMENT CHECK}\\
{[STEP 2 - MATCH]:}\\
Compare your predicted label with the given label.\\
\textit{CASE A} - Labels MATCH:\\
Confirm rationale is grounded in specific clause terms.\\
Proceed to finalize.\\
\textit{CASE B} - Labels DO NOT MATCH:\\
Identify what you missed or misinterpreted.\\
Unlearn your initial assumption.\\
Re-examine the clause strictly through the lens of the given label.\\
Rewrite the rationale to correctly reflect the given label.\\\\
\textbf{PHASE 3: FINALIZE}\\
{[STEP 3 - OUTPUT]:}\\
Output ONLY the final corrected rationale that is short, concise, and precise, starting with:\\
This clause is ambiguous. [your rationale here]\\
OR\\
This clause is not ambiguous. [your rationale here]\\
Do not output any intermediate steps, mismatches, or phase headings.\\
\textit{Clause:} []\\
\textit{Label: } []

\section{Prompting Templates for Baselines}
\label{AppendixA}
\subsection{Direct Prompting}
\label{AppendixA1}
\textbf{Instructions:} You are an expert in legal contract analysis. Classify the given contract clause as either "Ambiguous" or "Not Ambiguous" and provide a short, concise, and precise rationale.\\
\textbf{Input:}\\
\emph{Contract Clause:} []\\
\textbf{Output:}\\
\emph{Label:} [Ambiguous / Not Ambiguous]\\
\emph{Rationale:} [Explanation of the predicted label]
\subsection{Chain-of-Thought (CoT) Prompting}
\label{AppendixA1}
\textbf{Instruction:} You are an expert in legal contract analysis. Given the following contract clause, think step by step to classify it as Ambiguous or Not Ambiguous and provide a short, concise, and precise rationale.\\
\textbf{Input:}\\
\emph{Contract Clause:} []\\
\textbf{Output:}\\
Step 1: Identify legally significant keywords or phrases and evaluate if they introduce ambiguity\\
Step 2: If ambiguous, explain how the identified terms create conflicting interpretations; if not ambiguous, explain why the clause has a single clear interpretation\\
Step 3: If ambiguous, highlight the resulting legal consequences where applicable\\
\emph{Label:} [Ambiguous / Not Ambiguous]\\
\emph{Rationale:} [rationale behind the classified label]\\

\subsection{3-Shot Prompting}
\textbf{Instruction:} You are an expert in legal contract analysis. Given the following contract clause, classify it as Ambiguous or Not Ambiguous and provide a short, concise, and precise rationale.\\
\textbf{Definitions:}\\
\textbf{Ambiguous:} A clause that can be reasonably interpreted or constructed in more than one way, containing vagueness, incompleteness, or referential ambiguity.
\begin{itemize}
\item \textbf{Vagueness:} A clause that contains phrases allowing for borderline cases, leading to unclear boundaries of meaning.
\item \textbf{Incompleteness:} A grammatically correct clause that lacks enough information for a single, clear interpretation.
\item \textbf{Referential:} A word or phrase in a clause lacks a clear reference, leading to confusion about its meaning.
\end{itemize}
\textbf{Not Ambiguous:} A clause that has a single clear and unambiguous interpretation.\\
\textbf{Example 1:}\\
\emph{Contract Clause:} []\\
\emph{Label:} Ambiguous\\
\emph{Rationale:} []\\
\textbf{Example 2:}\\
\emph{Contract Clause:} []\\
\emph{Label:} Not Ambiguous\\
\emph{Rationale:} []\\
\textbf{Example 3:}\\
\emph{Contract Clause:} []\\
\emph{Label:} Ambiguous\\
\emph{Rationale:} []\\
\textbf{NOW YOUR TURN}\\
\textbf{Input:}\\
\emph{Contract Clause:} []\\
\textbf{Output:}\\
\emph{Label:} [Ambiguous / Not Ambiguous]\\
\emph{Rationale:} [rationale behind the classified label]
\subsection{Instruction Fine-tuning}
\label{AppendixA3}
\textbf{Instructions:} You are an expert in legal contract analysis. Classify the given contract clause as either "Ambiguous" or "Not Ambiguous".\\
\textbf{Input:}\\
\emph{Contract Clause:} []\\
\textbf{Output:}\\
\emph{Label:} [Ambiguous or Not Ambiguous]

\section{Models and Hyperparameters}
\label{AppendixB}
\textbf{\emph{Models:}} We use seven open-weight models, including encoder-only models BERT \cite{devlin2019bert} and RoBERTa \cite{liu2019roberta}; encoder-decoder models Flan-T5 \cite{chung2024scaling}; decoder-only models Qwen-2.5 \cite{qwen2} and Llama-3.2 \cite{touvronllama}; and domain-specific models Legal-BERT and Contracts-BERT\cite{chalkidis2020legal}, which serve as strong baselines for contract understanding \cite{singh2026evaluating}.

\textbf{\emph{Hyperparameters:}} For the proposed post-training approaches, all models are publicly available pre-trained models from Hugging Face. We evaluate different learning rates ($1\text{e-}4$, $2\text{e-}4$, $3\text{e-}4$, $1\text{e-}5$, $2\text{e-}5$, $3\text{e-}5$) and batch sizes ($4$, $8$, $16$, $32$) to identify the optimal settings for each model and approach via grid search. Post-training runs for a maximum of $10$ epochs. Experiments are conducted on a single NVIDIA V100 GPU. For supervised fine-Tuning, the batch size and learning rate are: $32$ and $3\text{e-}5$ for BERT; $32$ and $2\text{e-}5$ for RoBERTa; $32$ and $2\text{e-}4$ for Legal-BERT; $32$ and $1\text{e-}4$ for Contracts-BERT; $4$ and $3\text{e-}4$ for Flan-T5; $4$ and $2\text{e-}4$ for Qwen-2.5; and $4$ and $1\text{e-}4$ for Llama-3.2, with a maximum of $10$ epochs. For Instruction Fine-Tuning and PPI, the values are: $4$ and $1\text{e-}4$ for Flan-T5; $4$ and $1\text{e-}4$ for Qwen-2.5; and $4$ and $2\text{e-}4$ for Llama-3.2, with a maximum of $10$ epochs. 
For LAURA, the values are: $4$ and $3\text{e-}4$ for Flan-T5; $4$ and $2\text{e-}4$ for Qwen-2.5; and $4$ and $3\text{e-}4$ for Llama-3.2, with a maximum of $10$ epochs.
All models use the AdamW optimizer. The results are reported using a single seed set to 
$2018$ due to limited computational resources.

\section{LAURA Examples }
\label{AppendixC}
This section provides example of predicted labels and rationales using the LAURA + IRAC-Unlearning approach with the Flan-T5 model, including both correct and incorrect labels with their rationales, along with the ground truth and corresponding clause.
\subsection{Correct Predicted Labels with their Rationales}
\label{H1}
1. \textbf{Clause:} \emph{The place of the arbitration will be Chicago, Illinois.
}\\
\textbf{Ground Truth:} Not Ambiguous\\
\textbf{Prediction:} Not Ambiguous \\
\textbf{Rationale:} This clause is not ambiguous. The specific location of the arbitration venue is clearly stated, leaving no room for multiple interpretations.\\\\
2. \textbf{Clause:} \emph{Throughout the Term of the Agreement, Metavante shall maintain a Services Continuity Plan (the "Plan") in compliance with applicable regulatory requirements.
}\\
\textbf{Ground Truth:} Ambiguous\\
\textbf{Prediction:} Ambiguous \\
\textbf{Rationale:} This clause is ambiguous. The term "applicable regulatory requirements" is unclear as it does not specify which regulatory requirements apply, leading to potential uncertainty in interpretation.\\\\
3. \textbf{Clause:} \emph{ABG may, in ABG's sole discretion, allocate and apply payments it receives from PAPA JOHN'S hereunder.
}\\
\textbf{Ground Truth:} Ambiguous\\
\textbf{Prediction:} Ambiguous \\
\textbf{Rationale:} This clause is ambiguous. The term "sole discretion" lacks a clear, enforceable standard, leading to potential uncertainty in how ABG may allocate and apply payments.\\\\
4. \textbf{Clause:} \emph{Operator shall ensure that operation, service and maintenance records concerning Operator's activities hereunder are properly created and maintained at all times.
}\\
\textbf{Ground Truth:} Ambiguous\\
\textbf{Prediction:} Ambiguous \\
\textbf{Rationale:} The term "properly created and maintained" lacks specificity regarding the standards and conditions under which the records must be maintained, leading to potential disputes over compliance.
\subsection{Incorrect Predicted Labels with their Rationales}
\label{H2}
1. \textbf{Clause:} \emph{Neither Party shall be held liable for any failure to perform that is due to any cause or circumstance beyond the reasonable control of such Party, including without limitation a demand for such Products and other products manufactured by Company which exceeds Company's ability to supply them, earthquakes, fire, accidents, floods, storms, other Acts of God, riots, wars, rebellions, strikes, lockouts or other labor disturbances, national or international emergencies, failure to secure materials or equipment from usual sources of supply, failure of carriers to furnish transportation, government rules, regulations, acts, orders, world financial crisis, restrictions or requirements or any other cause or circumstance beyond the reasonable control of such Party.}\\
\textbf{Ground Truth:} Not Ambiguous\\
\textbf{Prediction:} Ambiguous \\
\textbf{Rationale:} This clause is ambiguous. The term "reasonable control" is subjective and lacks a clear standard for what constitutes reasonable control, leading to potential disputes over what constitutes reasonable control.\\\\
2. \textbf{Clause:} \emph{On or before the [***] after the Effective Date, AMAG shall provide Antares with an updated written [***] rolling forecast of the volume of Product, sample Product and Trainers that AMAG then anticipates will be required to be produced and delivered to AMAG during [***] (the "Forecast").}\\
\textbf{Ground Truth:} Not Ambiguous\\
\textbf{Prediction:} Ambiguous \\
\textbf{Rationale:} This clause is ambiguous. The term "[***] after the Effective Date" lacks specificity regarding the timing and conditions under which AMAG must provide a forecast, leading to potential uncertainty in obligations.\\\\
3. \textbf{Clause:} \emph{Neither Channel Partner, iPass, nor its suppliers warrant, nor do they assume responsibility for, any consequences suffered by any person as a result of Internet access including, without limitation, those suffered as a result of accessing Internet information and content, such as, by way of example only, the possibility of contracting computer viruses, accessing information with offensive, inaccurate or inappropriate content.}\\
\textbf{Ground Truth:} Ambiguous\\
\textbf{Prediction:} Not Ambiguous \\
\textbf{Rationale:} This clause is not ambiguous. The terms "warranty," "no liability," and "consequences" are clear and would be understood by a reasonable person to mean that neither the Channel Partner, iPass, nor its suppliers are liable for any consequences arising from Internet access.\\\\
4. \textbf{Clause:} \emph{By way of illustration, extending the use of the Licensed Product to patients of different age parameters who have multi-drug resistant HIV infection shall not be deemed a new indication distinct from the Initial Indication.}\\
\textbf{Ground Truth:} Ambiguous\\
\textbf{Prediction:} Not Ambiguous \\
\textbf{Rationale:} This clause is not ambiguous. The term "different age parameters" is clear in its intent to indicate that the use of the Licensed Product to patients of different age parameters does not constitute a new indication, leaving no room for multiple interpretations.

\section{Additional Analysis}
\label{AppendixF}
\textbf{In terms of rationale quality}, annotators find that LAURA's rationales for incorrectly labelled clauses are unconvincing and verifiably incorrect. For false positives, the model flags phrases that are clear within the clause context, such as \textit{reasonable control} in a force majeure clause where such terms carry a well-established legal meaning, and identifies redacted content \texttt{[***]} as ambiguous even when the ground truth is not ambiguous (Examples 1 and 2, Appendix \ref{AppendixC}). For false negatives, LAURA selects random phrases and provides unconvincing rationales to justify its predictions (Examples 3 and 4, Appendix \ref{AppendixC}). Nevertheless, as identification performance improves, the rationale quality is expected to improve correspondingly, making it increasingly useful for legal practitioners.

\end{document}